\def\year{2018}\relax
\documentclass[letterpaper]{article} 
\pdfoutput=1
\usepackage{aaai18}  
\usepackage{times}  
\usepackage{helvet}  
\usepackage{amssymb}  
\usepackage{subcaption}
\usepackage{courier}  
\usepackage{url}  
\usepackage{graphicx}  
\usepackage{verbatim}
\usepackage[table]{xcolor}
\title{Recurrent Neural Network Language Models for\\Open Vocabulary Event-Level Cyber Anomaly Detection}

\author{Aaron Tuor,\textsuperscript{1}
Ryan Baerwolf,\textsuperscript{2}
Nicolas Knowles,\textsuperscript{2}\\
{\bf \Large Brian Hutchinson,\textsuperscript{1,2}
Nicole Nichols\textsuperscript{1} \and
Robert Jasper\textsuperscript{1}}\\
\textsuperscript{1}Pacific Northwest National Laboratory\\Richland, Washington\\
\textsuperscript{2}Western Washington University\\Bellingham, Washington}

\begin{document}
\maketitle

\begin{abstract} 
 	Automated analysis methods are crucial aids for monitoring and defending a network to protect the sensitive or confidential data it hosts.  
    This work introduces a flexible, powerful, and unsupervised approach to detecting anomalous behavior in computer and network logs; 
    one that largely eliminates domain-dependent feature engineering employed by existing methods.  
    By treating system logs as threads of interleaved ``sentences'' (event log lines) to train online unsupervised neural network language models, our approach provides an adaptive model of normal network behavior.
    We compare the effectiveness of both standard and bidirectional recurrent neural network language models at detecting malicious activity within network log data.
    Extending these models, we introduce a tiered recurrent architecture, which provides context by modeling sequences of users' actions over time.
    Compared to Isolation Forest and Principal Components Analysis, two popular anomaly detection algorithms, we observe superior performance on the Los Alamos National Laboratory Cyber Security dataset. 
    For log-line-level red team detection, our best performing character-based model provides test set area under the receiver operator characteristic curve of 0.98, demonstrating the strong fine-grained anomaly detection performance of this approach on open vocabulary logging sources.
\end{abstract}

\section{Introduction}

To minimize cyber security risks, it is essential that organizations be able to rapidly detect and mitigate malicious activity on their computer networks.
These threats can originate from a variety of sources including malware, phishing, port scanning, etc.  
Attacks can lead to unauthorized network access to perpetrate further damage such as theft of credentials, intellectual property, and other business sensitive information.
In a typical scenario, cyber defenders and network administrators are tasked with sifting through vast amounts of data from various logging sources to assess potential security risks.
Unfortunately, the amount of data for even a modestly-sized network can quickly grow beyond the ability of a single person or team to assess, leading to delayed response. 
The desire for automated assistance has and continues to encourage inter-domain research in cyber security and machine learning.

Signature-based approaches for automated detection can be highly effective for characterizing individual threats.   
Despite their high precision, they suffer from low recall and may fail to detect subtle mutations or novel attacks.
Alternatively, given an unlabeled training set of typically benign activity logs, one can build a model of ``normal behavior''. 
During online joint training and evaluation of this model, patterns of normal usage will be reinforced and {\em atypical} malicious activity will stand out as anomalous.
The features used to identify unusual behavior are typically statistical feature vectors associated with time slices,
e.g., vectors of counts for types of activities taking place in a 24-hour window. Such systems developed in research have been criticized as brittle to differences in site-specific properties of real-world operational networks such as security constraints and variable usage patterns \cite{Sommer2010outside}. 

The approach we introduce aims to minimize site-specific assumptions implicit in feature engineering, and effectively model variability in network usage by direct online learning of language models over log lines.  
Language models assign probabilities to sequences of tokens 
and are a core component of speech recognition, machine translation, and other language processing systems. 
Specifically, we explore the effectiveness of several recurrent neural network (RNN) language models for use in a network anomaly detection system. 
Our system dynamically updates the network language model each day based on the previous day's events.
When the language model assigns a low probability to a log-line it is flagged as anomalous.  
There are several advantages to this approach:

\begin{enumerate}
    \item {\bf Reduced feature engineering:} Our model acts directly on raw string tokens, rather than hand-designed domain-specific statistics.  
        This dramatically reduces the time to deployment, and makes it agnostic to the specific network or logging source configuration.  
        It also removes the ``blind spots'' introduced when tens of thousands of log-lines are distilled down to a single aggregated feature vector, 
        allowing our model to capture patterns that would have otherwise been lost.
    \item {\bf Fine grained assessment:} The response time for analysts can be improved by providing more specific and relevant events of interest. Baseline systems that alert to a user's day aggregate require sifting through tens of thousands of actions.
        Our approach can provide log-line-level or even token-level scores to the analyst, helping them quickly locate the suspicious activity.
        \item {\bf Real time processing:}
        With the ability to process events in real time and fixed bounds on memory usage which do not grow over time, our approach is suitable for the common scenario in which log-line events are appearing in a high-volume, high-velocity log stream.
\end{enumerate}
         
 We assess our models using the publicly available Los Alamos National Laboratory (LANL) Cyber Security Dataset, 
which contains real (de-identified) data with ground truth red team attacks, and demonstrate 
 language models definitively outperforming standard unsupervised anomaly detection approaches.

\section{Prior work} 

Machine learning has been widely explored for network anomaly detection, 
with techniques such as isolation forest \cite{gavai2015supervised,LiuTingZhou08} 
and principal component analysis \cite{novakov2013studies,Ringberg2007SensitivityOP} attracting significant interest. 
Machine learning classifiers ranging from decision trees to Na\"{\i}ve Bayes have been used for cyber security tasks such as malware detection, network intrusion, and insider threat detection. 
Extensive discussion of machine learning applications in cyber security is presented in 
\cite{bhattacharyya2013network,buczak2016survey,dua2016data,kumar2010use,zuech2015intrusion,rubin2016anomaly}.

Deep learning approaches are also gaining adoption for specialized cyber defense tasks. 
In an early use of recurrent neural networks, Debar, Becker, and Siboni \shortcite{debar1992neural} model sequences of Unix shell commands for network intrusion detection.
Anomaly detection has been demonstrated using deep belief networks on the KDD Cup 1999 dataset \cite{alrawashdeh2016toward}, 
and Bivens et al. \shortcite{bivens2002network} use multi-layer perceptrons for the DARPA 1999 dataset. 
Both approaches use aggregated features and synthetic network data.
 Tuor et al. \shortcite{tuor2017deep} and Veeramachaneni et al. \shortcite{veeramachaneniAI2} both employ deep neural network autoencoders for unsupervised network anomaly detection using time aggregated statistics as features.
  
Some works of note have been previously published on the LANL data. 
Turcotte, Heard and Kent \shortcite{turcotte2016modelling} develop an online statistical model for anomaly detection in network activity using Multinomial-Dirichlet models. 
Similarly, Turcotte et al. \shortcite{turcotte2016poisson} use Poission Factorization \cite{gopalan2013scalable} on the LANL authentication logs. 
A user/computer authentication count matrix is constructed by assuming each count comes from a Poisson distribution parameterized by latent factors for users and computers. 
The learned distributions are then used to predict unlikely authentication behavior. 

Several variants of tiered recurrent networks have been explored in the machine learning and natural language processing communities \cite{koutnik2014clockwork,ling2015character,ling2015finding,chung2015gated}. 
They are often realized by a lower tier pre-processing network, whose output is fed to an upper tier network and the separate tiers are jointly trained. 
Ling et al. \shortcite{ling2015character} use a character-level convolutional neural network to feed a word level long short-term memory (LSTM) RNN for machine translation, with predictions made at the word-level. 
Both Hwang and Sung \shortcite{hwang2016character} and Ling et al. \shortcite{ling2015finding} use a character-based LSTM to feed a second word or utterance-based LSTM for language modeling. 
Pascanu et al. \shortcite{pascanu2015malware} create activity models from real world data on a per-event (command) basis and sequences of system calls are then modeled using RNN and echo state networks. 
The learned features are used to independently train neural network and logistic regression classifiers. 
Max pooling is applied to hidden layers of the unsupervised RNN for each time step in a session and the result is concatenated to the final hidden state to produce feature vectors for the classifier. 
This is similar to our tiered approach, in which we use the average of all hidden states concatenated with the final hidden state as input to the upper-tier RNN. 
In contrast, our model is completely unsupervised and all components are jointly trained. 

\section{Approach} 

Our approach learns {\it normal} behavior for users, processing a stream of computer and network log-lines as follows:
\begin{enumerate}
    \item \setlength{\parskip}{-3pt} Initialize model weights randomly
    \item For each day $k$ in chronological order:
        \begin{enumerate}
            \item Given model $M_{k-1}$, produce log-line-level anomaly scores for all events in day $k$
            \item Optionally, produce an aggregated anomaly score each user for day $k$ (from the log-line-level scores)
            \item Send per-user-day or per-user-event anomaly scores in rank order to analysts for inspection
            \item Update model weights to minimize loss on all log-lines in day $k$, yielding model $M_k$
        \end{enumerate}
\end{enumerate}

This methodology interleaves detection and training in an online fashion.  
In this section we detail the components of our approach.

\subsection{Log-Line Tokenization} \label{subsec:token}
To work directly from arbitrary log formats, we treat log-lines as sequences of tokens. 
For this work, we consider two tokenization granularities: word-level and character-level.  

For word tokenization, we assume that tokens in the log-line are delimited by a known character (e.g., space or comma).  
After splitting the log-lines on this delimiter, we define a shared vocabulary of ``words'' over all log fields, consisting of the sufficiently-frequent tokens appearing in the training set.
To allow our model to handle previously unseen tokens, we add an ``out of vocbulary'' token to our vocabulary, \verb|<oov>|. 
(For instance, not every IP address will be represented in a training set; likewise, new PCs and users are continually being added to large networks.) 
To ensure that \verb|<oov>| has non-zero probability, we replace sufficiently infrequent tokens in the training data with \verb|<oov>|.  
During evaluation, tokens not seen before are labeled \verb|<oov>|.
In order to accommodate shifting word distributions in an online environment, a fixed size vocabulary could be periodically updated using a sliding window of word frequency statistics. For simplicity, we assume we have a fixed training set from which we produce a fixed vocabulary.

To avoid the challenges of managing a word-level vocabulary, we also develop language models using a character-level tokenization. 
In this case our primitive vocabulary, the alphabet of printable ASCII characters, circumvents the open vocabulary issue by its ability to represent any log entry irrespective of the network, logging source, or log field.  With character-level tokenization, we keep the delimiter token in the sequence, to provide our models with cues to transitions between log-line fields.

\subsection{Recurrent Neural Network Language Models} \label{subsec:rnns}
To produce log-line-level anomaly scores, we use recurrent neural networks in two ways: 
1) as a language model over individual log-lines, and 
2) to model the state of a user over time.  
We first present two recurrent models that focus only on (1), 
and then a tiered model that accomplishes both (1) and (2). 
Both were implemented\footnote{Code will soon be available at https://github.com/pnnl/safekit} for our experiments using TensorFlow \cite{tensorflow2015-whitepaper}.

\subsubsection{Event Model (\texttt{EM}).}\label{subsec:simplemodel}
First we consider a simple RNN model that operates on the token (e.g., word) sequences of individual log-lines (events). 
Specifically, we consider a Long Short-Term Memory (LSTM) \cite{hochreiter1997long} 
network whose inputs are token embeddings and from whose output we predict distributions over the next token.

For a log-line with $K$ tokens, each drawn from a shared vocabulary of size $C$, let $\mathcal{X}_{(1:K)} = {\bf x}_{(1)}, {\bf x}_{(2)}, \dots, {\bf x}_{(K)}$ denote a sequence of one-hot representations of the tokens (each ${\bf x}_{(t)} \in \mathbb{R}^C$).  

In this model, the hidden representation at token $t$, $\textbf{h}_{(t)}$, from which we make our predictions, is a function of ${\bf x}_{(1)}, {\bf x}_{(2)}, \dots, {\bf x}_{(t)}$ according to the usual LSTM equations:

{\small
\begin{eqnarray}
    {\bf h}_{(t)} & = & {\bf o}_{(t)} \circ \tanh({\bf c}_{(t)})\\ 
    {\bf c}_{(t)} & = & {\bf f}_{(t)} \circ {\bf c}_{(t-1)} + {\bf i}_{(t)} \circ {\bf g}_{(t)}\\
    {\bf g}_{(t)} & = & \tanh\left( {\bf x}_{(t)} {\bf W}_{(g,x)} + {\bf h}_{(t-1)} {\bf W}_{(g ,h)}+ {\bf b}_{(g)} \right)\\
    {\bf f}_{(t)} & = & \sigma\left({\bf x}_{(t)} {\bf W}_{(f,x)} + {\bf h}_{(t-1)}{\bf W}_{(f ,h)} + {\bf b}_{(f)} \right)\\
    {\bf i}_{(t)} & = & \sigma\left({\bf x}_{(t)} {\bf W}_{(i,x)} + {\bf h}_{(t-1)} {\bf W}_{(i ,h)}+ {\bf b}_{(i)} \right)\\
    {\bf o}_{(t)} & = & \sigma\left({\bf x}_{(t)} {\bf W}_{(o,x)} + {\bf h}_{(t-1)}{\bf W}_{(o ,h)} + {\bf b}_{(o)} \right),
    \label{eqn:lstm}
\end{eqnarray}}
where the initial hidden and cell states, ${\bf c}_{(0)}$ and ${\bf h}_{(0)}$, are set to zero vectors, and 
 $\circ$ and $\sigma$ denote element-wise multiplication and logistic sigmoid, respectively.  
 Vector ${\bf g}_{(t)}$ is a hidden representation based on the current input and previous hidden state, while vectors ${\bf f}_{(t)}$, ${\bf i}_{(t)}$, and ${\bf o}_{(t)}$, are the standard LSTM gates.
The matrices (${\bf W}$) and bias vectors (${\bf b}$) are the model parameters.
We use each ${\bf h}_{(t-1)}$ to produce a probability distribution ${\bf p}_{(t)} $ over the token at time $t$, as follows:
\begin{equation}
    {\bf p}_{(t)}  = \mbox{softmax}\left({\bf h}_{(t-1)} {\bf W}_{(p)} + {\bf b}_{(p)}\right)
\end{equation}
We use cross-entropy loss,
\begin{equation}
    \frac{1}{K} \sum_{t=1}^{K} H({\bf x}_{(t)}, {\bf p}_{(t)}),
\end{equation}
for two important purposes: first, as per-log-line anomaly score and second, as the training objective to update model weights.  We train this model using stochastic mini-batch (non-truncated) back-propagation through time.

\begin{figure}
\centering
\includegraphics[width=.45\textwidth]{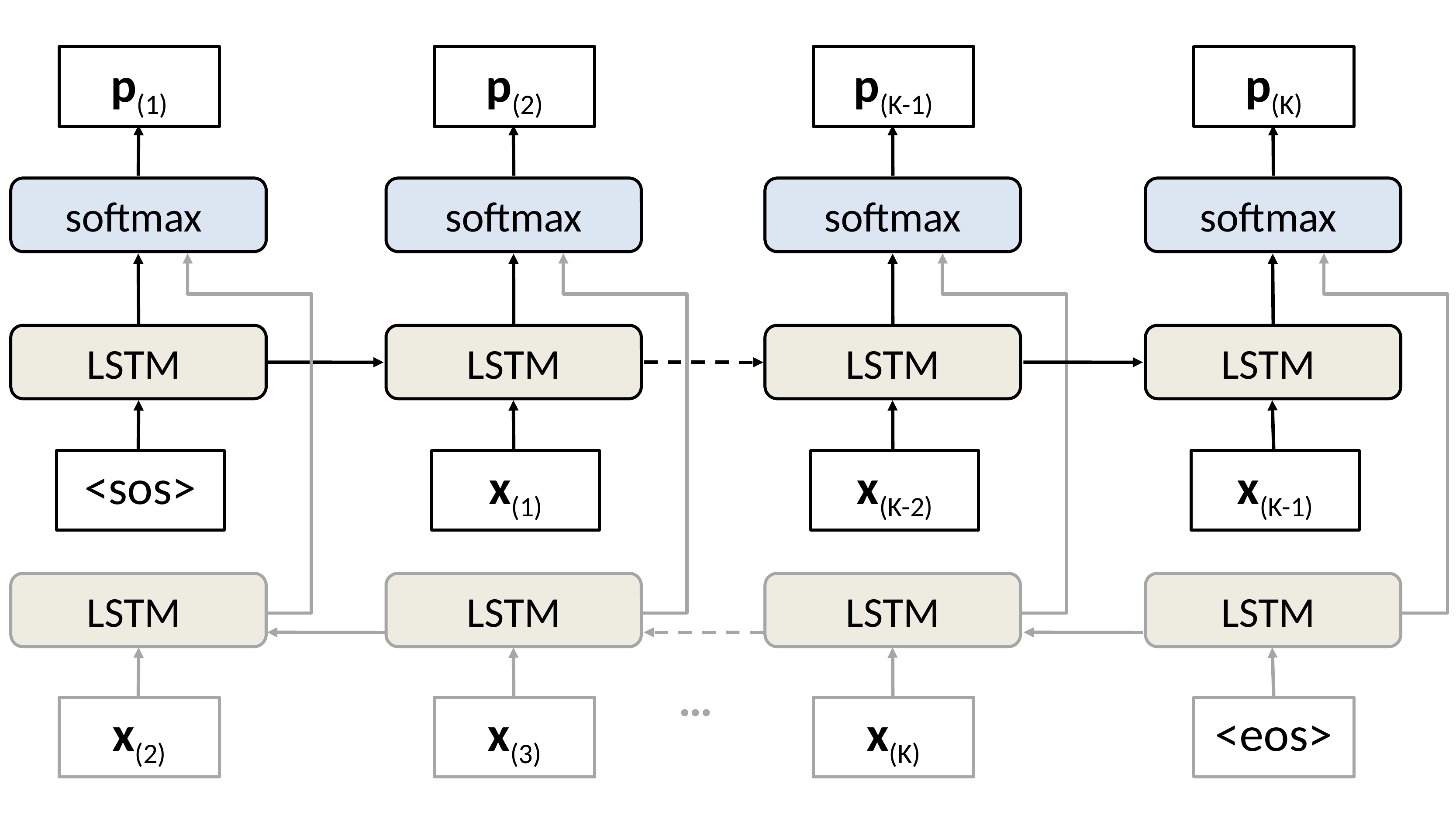}
\caption{Event Models. Set of black bordered nodes and connections illustrate the \texttt{EM} model while set of all nodes and connections illustrate the \texttt{BEM} model. }\label{fig:charlm}
\end{figure}

\subsubsection{Bidirectional Event Model (\texttt{BEM}).}\label{subsec:bidirmodel}
Following the language model formulation suggested in \cite{schuster1997bidirectional}, we alternatively model the structure of log lines with a {\it bidirectional} LSTM.
 We define a new set of hidden vectors $\mathbf{h}^b_{(K+1)}, \mathbf{h}^b_{(K)},\dots,\mathbf{h}^b_{(1)}$ by running the LSTM equations backwards in time (starting with initial zero cell and hidden states at time $K+1$ set to zero).  The weights $\mathbf{W}$ and biases $\mathbf{b}$ for the backward LSTM are denoted with superscript $b$.

 The probability distribution ${\bf p}_{(t)} $ over the token at time $t$ is then:
\begin{equation}
    {\bf p}_{(t)}  = \mbox{softmax}\left({\bf h}_{(t-1)} {\bf W}_{(p)} + {\bf h}^b_{(t+1)} {\bf W}_{(p)}^{b}+ {\bf b}_{(p)}\right)
\end{equation}

\subsubsection{Tiered Event Models (\texttt{T-EM}, \texttt{T-BEM}).}\label{subsec:tieredmodel}
\begin{figure}
\centering
\includegraphics[width=.47\textwidth]{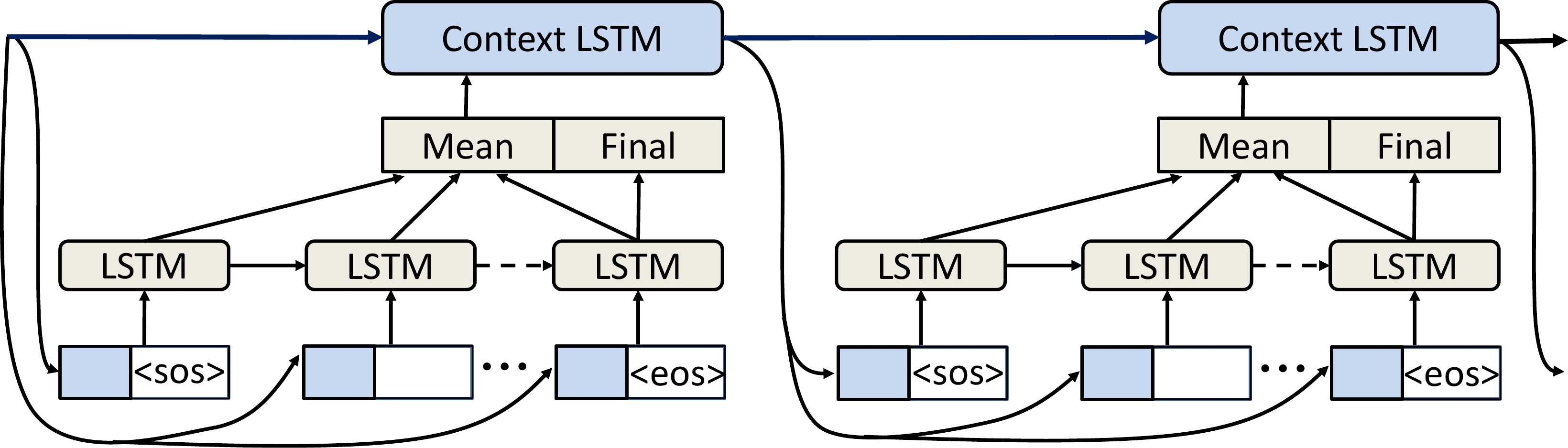}
\caption{Tiered Event Model (T-EM)} \label{fig:nlm}
\end{figure}

To incorporate inter-log-line context, we propose a two-tiered recurrent neural network.  
The lower-tier can be either event model (\texttt{EM} or \texttt{BEM}), but with the additional input of a context vector (generated by the upper-tier) concatenated to the token embedding at each time step.
The input to the upper-tier model is the hidden states of the lower-tier model.  This upper tier models the dynamics of user behavior over time, producing the context vectors provided to the lower-tier RNN. This model is illustrated in Fig. \ref{fig:nlm}.  

In this model, $x^{(u,j)}$ denotes user $u$'s $j$th log line, which consists of a sequence of tokens as described in the previous subsections.  The upper-tier models a sequence of user log lines, $x^{(u,1)}, x^{(u,2)},\dots,x^{(u,T_u)}$, using an LSTM.  
For each user $u$ and each log line $j$ in the user's log line sequence, a lower-tier LSTM is applied to the tokens of $x^{(u,j)}$.
The input to the upper-tier model at log-line $j$ is the concatenation of: 1) the final lower-tier hidden state(s) 
and 2) the average of the lower-tier hidden states.
In the case of a lower-tier EM, (1) refers to the hidden state at time $K$; for the BEM, (1) is the concatenation of the forward hidden state at time $K$ and the backward hidden state at time $1$.
For (2), we average over hidden states primarily to provide many short-cut connections in the LSTM, which aids trainability.
The output of the upper-tier LSTM at log-line $j$ is a hidden state ${\bf \hat{h}}^{(u,j)}$.  This hidden vector serves to provide context for the lower-tier model at the next time step: specifically, ${\bf \hat{h}}^{(u,j-1)}$ is concatenated to each of the inputs of the lower-tier model operating on the $j$th log-line.  Note that the upper-tier model serves only to propagate context information across individual log-lines;
no loss is computed directly on the values produced by the upper-tier model.  

The upper- and lower-tier models are trained jointly to minimize the cross-entropy loss of the lower-tier model.  We unroll the two-tier model for a fixed number of log-lines, fully unrolling each of the lower-tier models within that window.  The lower-tier model's cross-entropy loss is also used to detect anomalous behavior, as is described further in Section \ref{subsec:detectinganomaly}.

Minibatching becomes more challenging for the tiered model, as the number of log-lines per day can vary dramatically between users.  
This poses two problems: first, it introduces the possibility that the most active users may have a disproportionate impact on model weights; second, it means that toward the end of the day, there may not be enough users to fill the minibatch.  
To counteract the first problem, we fix the number of log-lines per user per day that the model will train on. The remaining log-lines are not used in any gradient updates.  
We leave compensating for the inefficiency that results from the second to future work.

\subsection{Baselines} \label{subsec:baselines}
Anomaly detection in streaming network logs often relies upon computing statistics over windows of time
and applying anomaly detection techniques to those vectors.  
Below we describe the aggregate features and two anomaly detection techniques that are typical of prior work. 

\subsubsection{Aggregate Features}
We first define the set of per-user-day features, which summarize users' activities in the day.
To aggregate the features that have a small number of distinct values (e.g. success/failure, logon orientation) 
we count the number of occurrences for each distinct value for the user-day. 
For fields that have a larger number of distinct values (pcs, users, domains), we count the number of common and uncommon events that occurred, rather than the number of occurrences of each distinct value (this approach avoids high dimensional sparse features). 
Furthermore, we define two categories of common/uncommon; to the individual entity/user, and relative to all users. 
A value is defined as uncommon for the user if it accounts for fewer than 5\% of the values observed in that field (up to that point in time), and common otherwise. A value is defined as uncommon for all users if it occurs fewer times than the average value for the field, and common otherwise.

For the LANL dataset, the prior featurization strategy yields a 108-dimensional aggregate feature vector per user-day.  
These feature vectors then serve as the input to the baseline models described next.

\subsubsection{Models}
We consider two baseline models.  
The first uses Principal Components Analysis (\verb|pca|) to learn a low dimensional representation of the aggregate features; 
the anomaly score is proportional to the reconstruction error after mapping the compressed representation back into the original dimension \cite{ShyuChenetal03}.  
The second is an isolation forest (\verb|iso|) based approach \cite{LiuTingZhou08}  as implemented in scikit-learn's outlier detection tools \cite{scikit-learn}. This was noted as the best performing anomaly detection algorithm in the recent DARPA insider threat detection program, \cite{gavai2015supervised}.   

\section{Experiments}
In this section we describe experiments to evaluate the effectiveness of the proposed event modeling algorithms.  
\subsection{Data} 
 
The Los Alamos National Laboratory (LANL) Cyber Security Dataset \cite{kent2016cyber} consists of event logs from LANL's  internal computer network collected over a period of 58 consecutive days. The data set contains over one billion log-lines from  authentication, process, network flow, and DNS logging sources. Identifying fields (e.g., users, computers, and processes) have been anonymized.
 
The recorded network activities included both normal operational network activity as well as a series of red team activities that compromised account credentials over the first 30 days of data.
Information about known red team attack events is used only for evaluation; our approach is strictly unsupervised.

For the experiments presented in this paper, we rely only on the authentication event logs, whose fields and statistics are summarized in Figure \ref{tab:auth}.  We  filter these events to only those log-lines linked to an actual user, removing computer-computer interaction events. Events on weekends and holidays contain drastically different frequencies and distributions of activities.  In a real deployment a separate model would be trained for use on those days, but because no malicious events were in that data it was also withheld.  

\begin{figure}
\begin{subfigure}{0.47\textwidth}
\centering
    \begin{tabular}{|l|r|r|}\hline
         {\bf Field} & {\bf Example}  & {\bf \# unique labels} \\\hline
         {\bf time} & {1} & {5011198}  \\\hline
        {\bf source user} & {C625@DOM1}  & {80553}\\\hline
        {\bf dest. user} & {U147@DOM1} & {98563}  \\\hline
        {\bf source pc} & {C625}  & {16230} \\\hline
        {\bf dest. pc} & {C625}  & {15895} \\\hline
        {\bf auth. type} & {Negotiate} & {29} \\\hline
        {\bf logon type} & {Batch}  & {10} \\\hline
        {\bf auth. orient} & {LogOn}  & {7} \\\hline
        {\bf success} & {Success}  & {2} \\\hline
    \end{tabular}
    \caption{} \label{tab:auth}
\end{subfigure}
\begin{subfigure}{0.47\textwidth}
\centering
    \begin{tabular}{|l|r|r|}\hline
         {\bf } & {\bf Dev}  & {\bf Test} \\\hline
         {\bf Days} & {1-12} & {13-58}  \\\hline
        {\bf \# Events} & $133M$ & $918M$\\\hline
        {\bf \# Attacks} & 316 & 385  \\\hline
        {\bf \# User-days} & 57 & 79 \\\hline
    \end{tabular}
    \caption{} \label{tab:split}
\end{subfigure}
    \caption{Dataset statistics: (a) Authentication log fields and statistics and (b) dataset splits.}
\end{figure}

Table \ref{tab:split} has statistics of our data split; the first 12 days serve as the development set, while the remaining 18 days are the independent test set. 

\subsection{Assessment Granularity} \label{subsec:detectinganomaly}
Our model learns normal behavior and assigns relatively high loss to events that are unexpected.
A principal advantage of our approach is this ability to score the anomaly of individual events, allowing us to flag at the event-level or aggregate anomalies over any larger timescale. 

For this work, we consider two timescales.  First, we assess based on individual events; a list of events would be presented to the analyst, sorted  descending by anomaly score. Second, to facilitate comparison with traditional aggregation methods, we aggregate anomaly scores over all of a user's events for the day (specifically, taking the max), producing a single anomaly score per-user, per-day.  In this scenario, a list of user-days would be provided to the analyst, sorted descending by anomaly score.  We refer to this approach as \verb|max|, because the anomaly scores provided to the analyst are produced by taking the maximum score over the event scores in the window for that user (where event-level scoring is just taking the max over a singleton set of one event).

 In order to counter systematic offsets in users' anomaly scores for a day we also consider a simple normalization strategy, which we refer to as \verb|diff|, by which every raw score is first normalized by subtracting the user's average event-level anomaly score for the day. 
 
\subsection{Metrics} 
We consider two performance metrics.
First, we assess results using the standard area under the receiver operator characteristic  curve (AUC) which characterizes the trade-off in model detection performance between true positives and false positives, effectively sweeping through all possible analyst budgets.
False positives are detections that are not truly red team events, while true positives are detections that are. 
 
To quantify the proportion of the data the analyst must sift through to diagnose malicious behavior on the network, we use the Average Percentile (AP) metric.
Specifically, for each red team event or user-day, we note the percentile of its anomaly amongst all anomaly scores for the day.  
We then average these percentiles for all of the malicious events or user-days.
Note that if all true malicious events or user-days are flagged as the most anomalous on the respective days, then $\texttt{AP} \approx 100$, while if all malicious events or user-days are ranked as the least anomalous on their respective days, $\texttt{AP} \approx 0$.
For both AUC and AP, a higher score is better.

Our model hyperparameters were manually tuned to maximize AP for day-level \texttt{diff} scores on the development set.  No separate training set is needed as our approach is unsupervised and trained online.

\subsection{Results and Analysis}

\begin{table}
    \centering
    \begin{tabular}{|r|c|r|r|} \hline 
        {\bf Model} & {\bf Tokenization}  & {\bf AUC} & {\bf AP} \\\hline
        \verb|pca|   & -          & 0.754 & 73.9 \\\hline
        \verb|iso|   & -          & 0.763 & 75.0 \\\hline\hline
        \verb|EM|    & Word       & 0.802 &79.3 \\\hline
        \verb|BEM|   & Word       & {\bf 0.876} & {\bf 87.0}\\\hline
        \verb|T-EM|  & Word       & 0.782 &  77.5 \\\hline
        \verb|T-BEM| & Word       & 0.864 &  85.7 \\\hline\hline
        \verb|EM|    & Char       & 0.750  & 70.9\\\hline
        \verb|BEM|   & Char       &{\bf 0.843}  & {82.9}\\\hline
        \verb|T-EM|  & Char       & 0.772 & 76.2\\\hline
        \verb|T-BEM| & Char       & { 0.837} & {\bf 82.9}\\\hline
    \end{tabular}
    \caption{User-day granularity test set AUC and AP. Language model anomaly scores calculated with average user-day normalization (\texttt{diff}).} \label{tab:aucap}
\end{table}

\begin{table}
    \centering
    \begin{tabular}{|l|r|r|r|r|}
          \multicolumn{3}{c}{\hspace{60pt}LOG}&\multicolumn{2}{c}{DAY}\\
           \hline  &\cellcolor{gray}\textcolor{white}{\bf \texttt{diff}} &\cellcolor{black}\textcolor{white}{\bf \texttt{max}}&\cellcolor{gray}\textcolor{white}{\bf \texttt{diff}} & \cellcolor{black}\textcolor{white}{\bf \texttt{max}}\\\hline
            \verb|W EM |  &0.964     &0.932& 0.802 &    0.794 \\\hline
            \verb|W BEM|    & {\bf 0.974}    &0.895& 0.876 &  0.811 \\\hline
            \verb|W T-EM|&  0.959   &0.948 &  0.782&   0.803   \\\hline
            \verb|W T-BEM|  &  0.959   & 0.902& 0.864  &   0.838   \\\hline\hline
            \verb|C EM |&  0.940   & 0.935 &  0.751&    0.754  \\\hline
            \verb|C BEM| & 0.973    &{\bf 0.979} & 0.843 & 0.846    \\\hline
            \verb|C T-EM|&  0.859  & 0.927 &  0.772&   0.809   \\\hline
            \verb|C T-BEM| & 0.945   & 0.969& 0.837& 0.854      \\\hline
    \end{tabular}
    \caption{Comparison of AUC for day-level and log-line-level analysis with and without user-day normalization. Figures \ref{fig:word} and \ref{fig:char} provide a visualization of these results.} \label{tab:gran1}
\end{table}
 \begin{figure*}
\minipage{0.47\textwidth}
  \includegraphics[width=\linewidth]{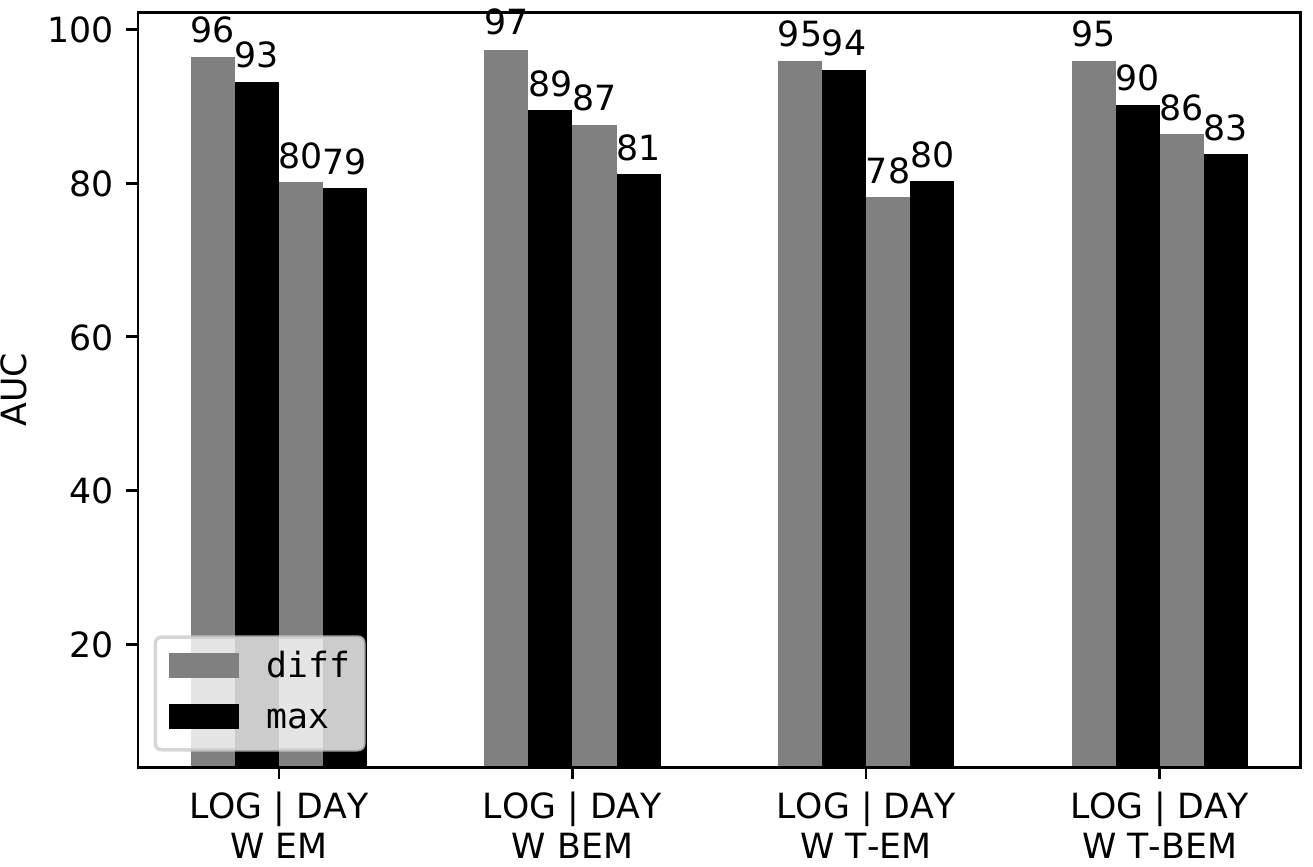}
\caption{Word model comparison of AUC at day-level and log-line-level granularities.} \label{fig:word}
\endminipage\hfill
\minipage{0.47\textwidth}
  \includegraphics[width=\linewidth]{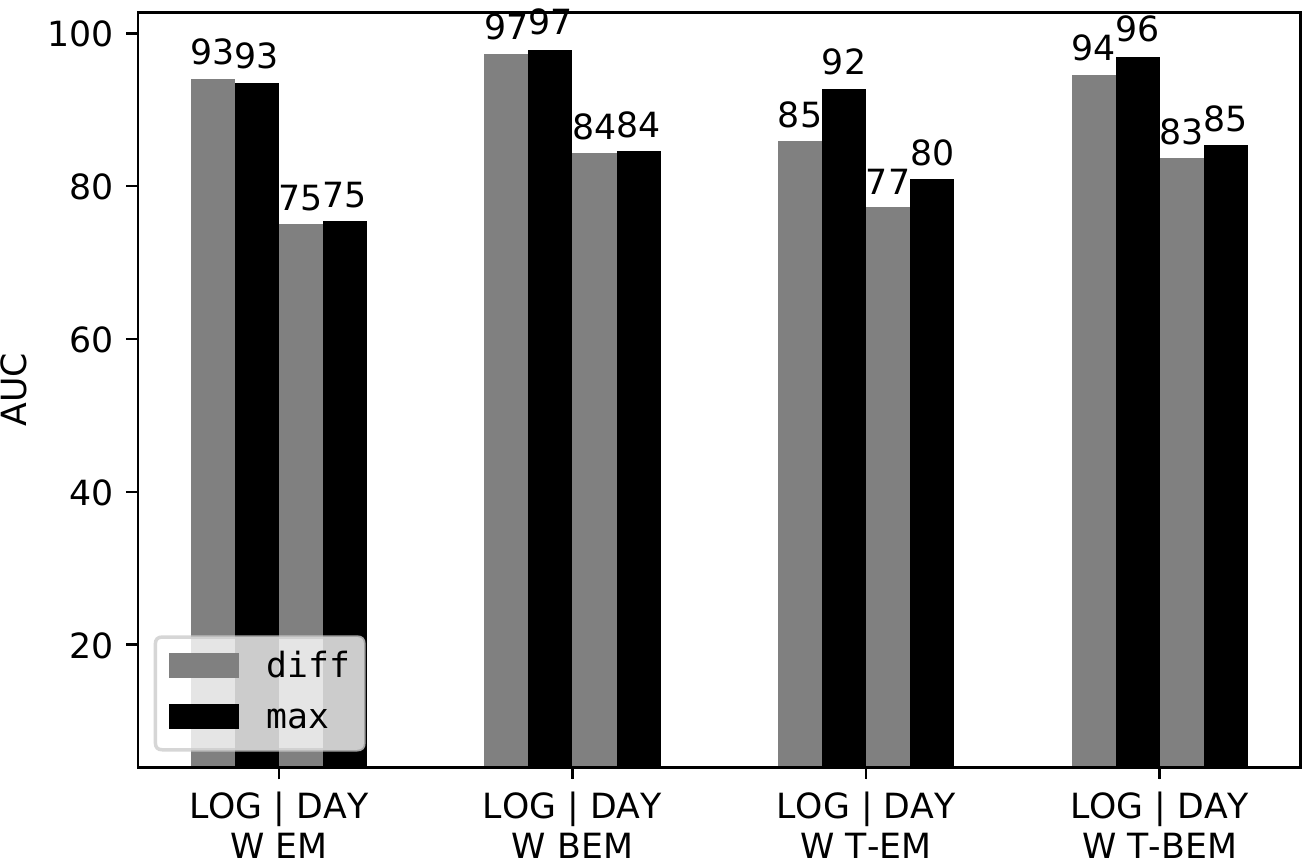}
\caption{Character model comparison of AUC at day-level and log-line-level granularities.} \label{fig:char}
\endminipage
\end{figure*}

We begin by exploring the user-day granularity performance.  Table \ref{tab:aucap} summarizes model detection performance at this granularity on the test set for the AUC and AP metrics using the {\verb diff } method to produce day level scores from the language models.   
A few trends are evident from these results.
First, the aggregate feature baselines have near-equivalent performance by both metrics, with the isolation forest approach having a slight edge.  We hypothesize the feature representation, which is common to these methods, could be a bottleneck in performance.  This highlights the ``blind spot'' issue feature engineering introduces.
Second, despite having only the context of a single log-line at a time, as opposed to features aggregated over an entire day, the event model (\verb|EM|) performs comparably to the baseline models when a forward pass LSTM network is used with a character tokenization and outperforms the baselines with word tokenization. The most pronounced performance gain results from using bidirectional models. Finally, word-level tokenization performs better than character-level; however, even the bidirectional {\it character} models perform appreciably better than the baselines. 

It is clear from these results that the tiered models perform comparably to, but not better than, the event-level models. This suggests that the event level model is able to characterize normal user behavior from information stored in the model weights of the network, which are trained each day to model user activity. Given the context of the past day's activity stored in the model weights, the categorical variables represented by the fields in an individual log line may eliminate the need for explicit event context modeling.
We leave tracking the state of individual computers, rather than users, to future work, but hypothesize that it may make the tiered approach more effective.

Next, we broaden our analysis of language modeling approaches, comparing performance across all language models, tokenization strategies, anomaly granularity, and normalization techniques. 
Figure \ref{fig:word} plots AUC for all language model types using word tokenization, contrasting \verb|max| and \verb|diff| normalization modes.  Figure \ref{fig:char} compares the same variations for character tokenization. Table \ref{tab:gran1} presents these results in tabular form.
With few exceptions, log-line-level granularity vastly outperforms day-level; this is true for both the character-level and word-level tokenization strategies, with an average gain of 0.1 AUC. The most interesting outcome of these comparisons is that word tokenization performance gains are heavily reliant on the \verb|diff| normalization, whereas for character tokenization the \verb|diff| normalization has a minor detrimental effect for some models. This suggests that the character-level model could be used to provide a more immediate response time, not having to wait until the day is done to obtain the day statistics used in \verb|diff| mode. The two tokenization strategies may in fact be complementary as the versatility and response time gains of a character tokenization come at the expense of easy interpretibility of a word tokenization: the word tokenization allows anomaly scores to be decomposed into individual log-line fields, enabling analysts to pinpoint features of the event that most contributed to it being flagged. Since we tuned hyper-parameters using \verb|diff| mode, the character-level model has potential to do better with additional tuning. 

 \begin{figure*}
\minipage{0.47\textwidth}
  \includegraphics[width=\linewidth]{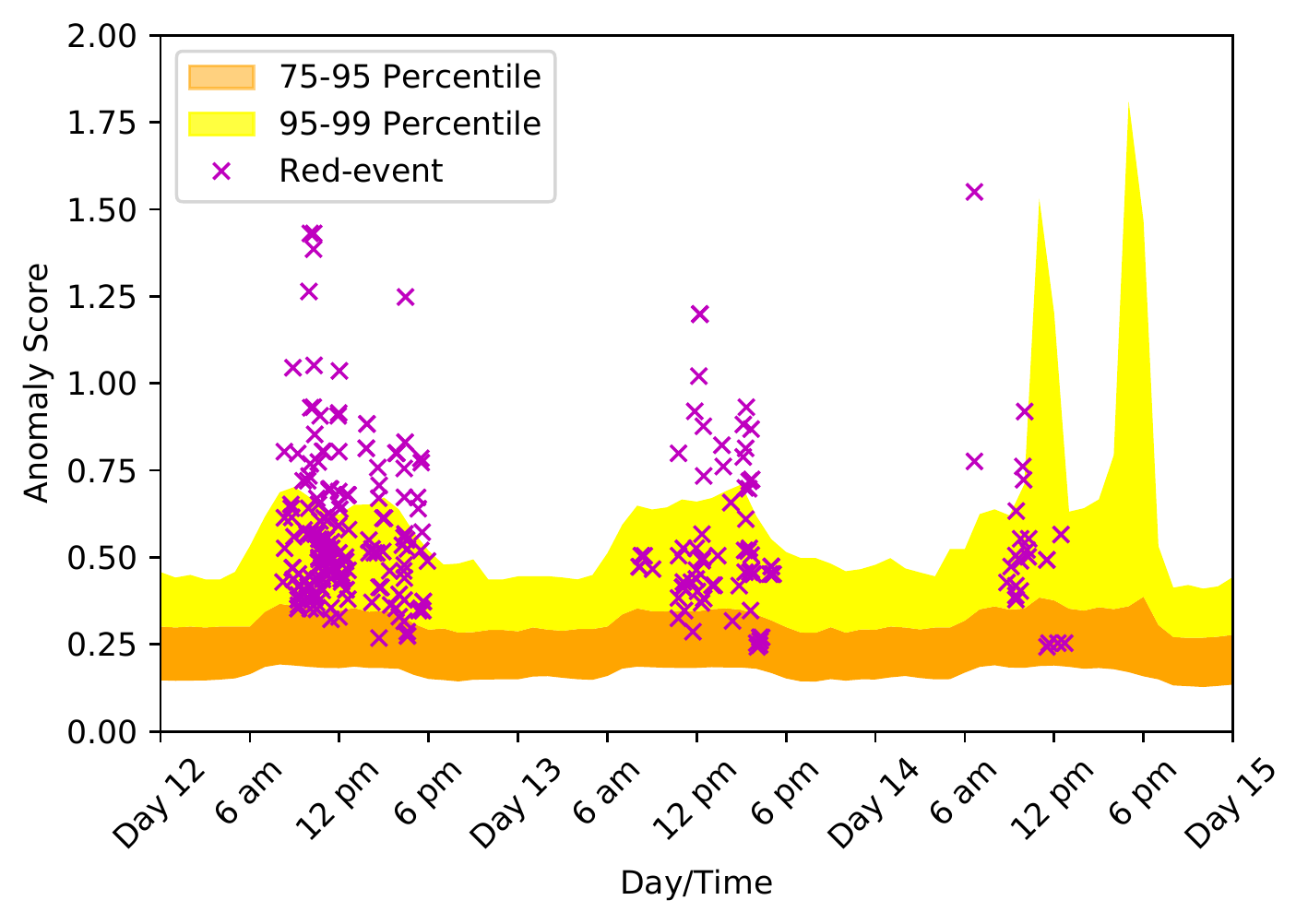}
\caption{Character-level red-team log-line anomaly scores in relation to percentiles over time.} \label{fig:char_perc}
\endminipage\hfill
\minipage{0.47\textwidth}%
 \includegraphics[width=\linewidth]{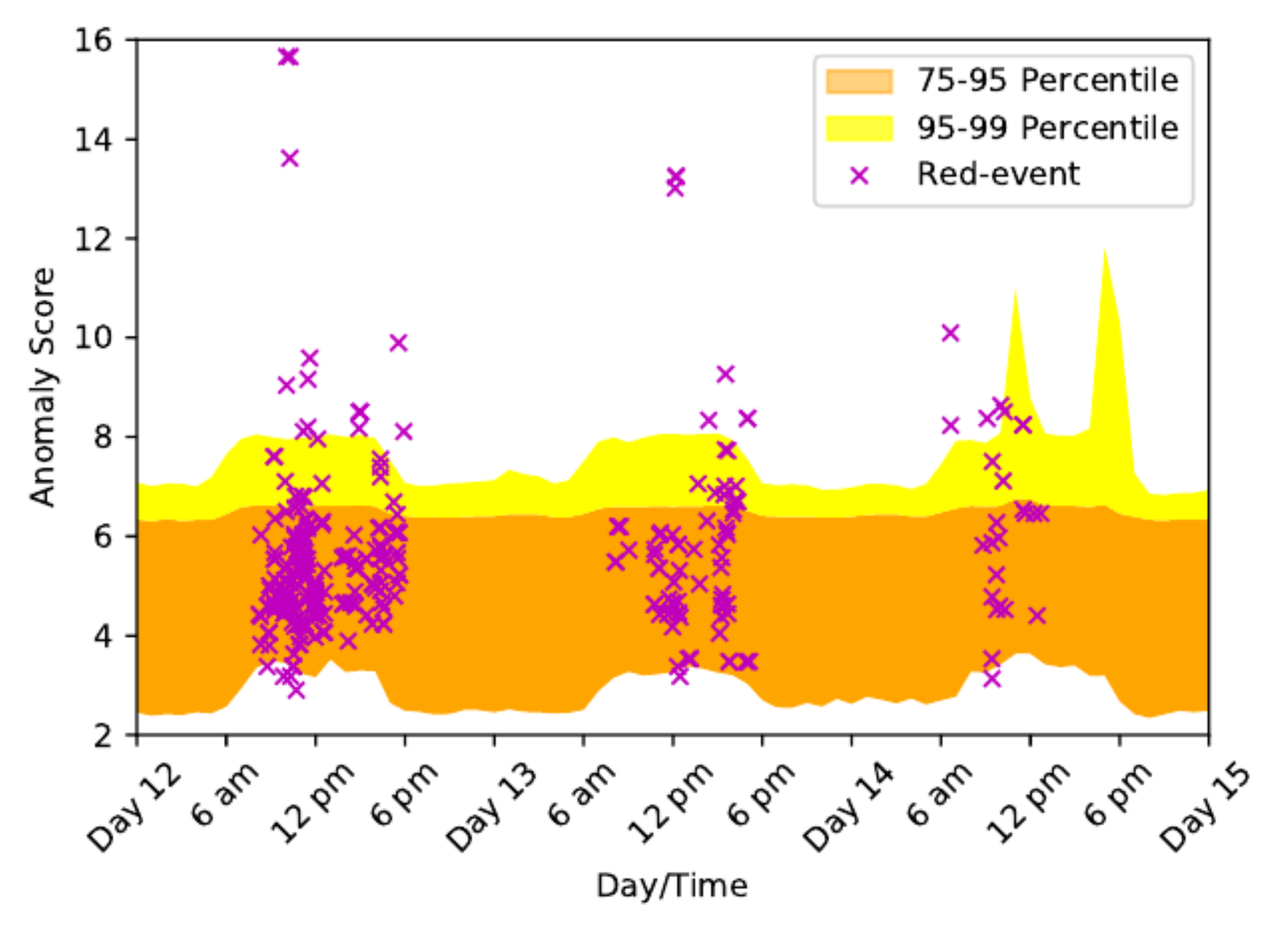}
\caption{Word-level red-team log-line anomaly scores in relation to percentiles over time.} \label{fig:word_perc}
\endminipage
\end{figure*}

Next, Figures \ref{fig:char_perc}  and  \ref{fig:word_perc} visualize the average percentiles of red team detections for the subset of the test set with the most red-team activity. Anomaly scores for both word and character tokenizations are computed {\em without} average user-day offset normalization. Red team log-line-level scores are plotted as purple x's with the $x$ coordinate being the second in time at which the event occurred and $y$ coordinate the anomaly score for that event. Percentile ranges are colored to provide context for the red-team anomaly scores against the backdrop of other network activity. The spread of non-normalized anomaly scores is much greater for the word-level tokenizations (Fig. \ref{fig:word_perc}) than character-level (Fig. \ref{fig:char_perc}), which could explain the different sensitivity of word level tokenization to normalization. Also notice that there is an expected bump in percentiles for windows of frequent red-team activity. Curiously, at the end of day 14 there are massive bumps for the 99th percentile, which suggest unplanned and un-annotated anomalous events on the LANL network for those hours. Notice that for the character tokenization almost all non-normalized red team anomaly scores are above the 95th percentile, with a large proportion above the 99th percentile. 

\begin{figure}
\includegraphics[width=.47\textwidth]{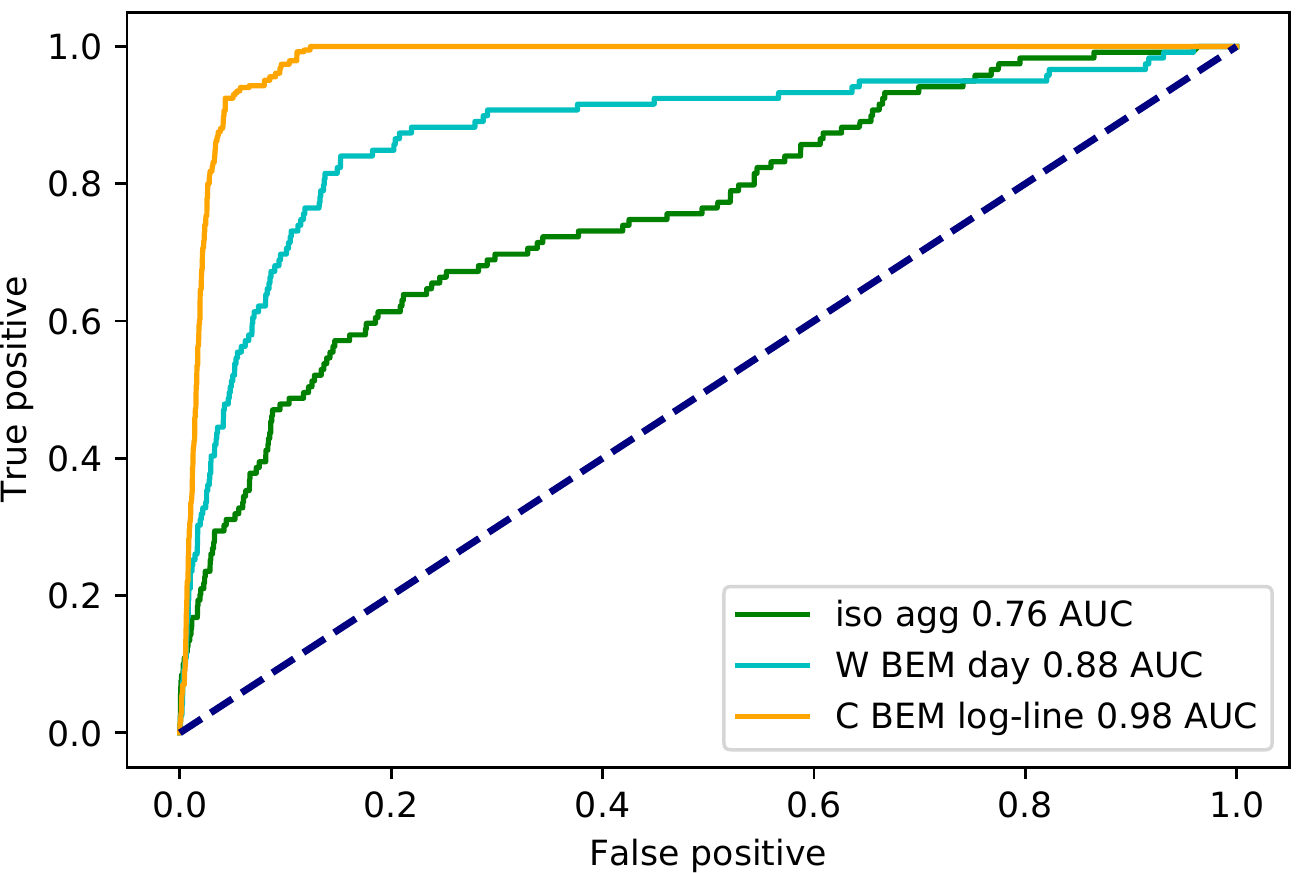}
\caption{ROC curves for best performing baseline, word language model evaluated at day-granularity, and character language model evaluated at log-line-granularity.} \label{fig:roc}
\end{figure}

Finally,
Figure \ref{fig:roc} plots the ROC curves for the best aggregate baseline (\verb|iso|), the best user-day granularity language model (word \verb|BEM|), and the best event-level granularity model (character \verb|BEM|). It illustrates the qualitatively different curves obtained with the baselines, the user-day granularity, and the event-level granularity. 

Since the proportion of red-team to normal events is vanishingly low in the data-set ($<0.001$\%), the false-positive rate is effectively the proportion of data flagged to achieve a particular recall. 
From this observation, Figure \ref{fig:roc} shows the character event model can achieve 100\% recall from only 12\% of the data whereas the other models considered only achieve 100\% recall when nearly all of the data has been handed to the analyst. Further, the character event model  can achieve 80\% recall by flagging only 3\% of the data whereas the word day language model needs 14\% of the data and the aggregate isolation forest model needs 55\% of the data to achieve the same result. 

\section{Conclusion}
This work builds upon advances in language modeling to address computer security log analysis, proposing an unsupervised, online anomaly detection approach. We eliminate the usual effort-intensive feature engineering stage, making our approach fast to deploy and agnostic to the system configuration and monitoring tools. It further confers the key advantage of event-level detection which allows for a near immediate alert response following anomalous activity. 

In experiments using the Los Alamos National Laboratory Cyber Security Dataset, bidirectional language models significantly outperformed standard methods at day-level detection. The best log-line-level detection performance was achieved with a bidirectional character-based language model, obtaining a 0.98 area under the ROC curve, 
showing  that for the constrained language domain of network logs, character based language modeling can achieve comparable accuracy to word based modeling for event level detection. We have therefore demonstrated a simple and effective approach to modeling dynamic networks with open vocabulary logs (e.g. with new users, PCs, or IP addresses).  

We propose to extend this work in several ways. 
First, potential modeling advantages of tiered architectures merit further investigation. The use of tiered architectures to track PCs instead of network users, or from a richer set of logging sources other than simply authentication logs may take better advantage of their modeling power. 
Next, we anticipate interpretability can become lost with such detailed granularity provided by log-line-level detection from a character-based model, therefore future work will explore alternate methods of providing context to an analyst.  
Finally, we are interested in exploring the robustness of this approach to adversarial tampering. Similarly performing models could have different levels of resilience that would lead to selection of one over another.

\subsubsection*{Acknowledgments}
The research described in this paper is part of the Analysis in Motion Initiative at Pacific Northwest National Laboratory. It was conducted under the Laboratory Directed Research and Development Program at PNNL, a multi-program national laboratory operated by Battelle for the U.S. Department of Energy.  The authors would also like to thank the Nvidia corporation for their donations of Titan X and Titan Xp GPUs used in this research.  

\bibliography{tuor}
\bibliographystyle{aaai}

\end{document}